%% file: main.tex
\documentclass[letterpaper, 10pt, journal, twoside]{IEEEtran}

\usepackage{amsmath,amssymb}
\usepackage{newtxtext,newtxmath}
\usepackage{graphicx}
\usepackage{booktabs}
\usepackage{cite}
\usepackage[hidelinks]{hyperref}

\begin{document}

\title{Sybil-Resilient Multi-Drone Exploration\\ of Indoor and Confined Spaces}

\author{Shyaam Mahesh, Aditya B Prakash, Rubal Sagwal, and Avinash Gautam%
\thanks{The simulator, the worlds and the result of every mission are available at \url{https://github.com/IDZTPHKKKS/Sybil4Drones}.}}

\maketitle

\begin{abstract}
When a team of drones explores a building together, each drone tells the others which frontier it is going to and shares the map it has built. This avoids duplicated effort, but it also means that the team trusts every identity from which it receives messages. In a Sybil attack, one compromised drone creates fake identities, and these fake teammates claim frontiers and send fabricated scans, so the real drones leave parts of the building unexplored and end up with walls in their maps that do not exist. We study five such attackers, including one that seals off the openings between floors, and five levels of defence that need no security infrastructure, only the drones' own sensors and maps: an audit of received map data against what a drone sees itself, a check that a body is present where an identity claims to be, trust only for identities whose body has been seen, a quarantine for identities that no drone has observed, and a rule that binds each observed body to one identity. Tests in simulation cover 70 maps of seven kinds, from offices and tunnels to three-storey buildings and caves. Without a defence, one compromised drone leaves on average up to 25\% of a map unexplored (45\% in offices), and the floor-sealing attacker cuts the explored part of a three-storey building to 82.3\%, or 34.6\% on the worst map. The quarantine brings exploration back to the level of an unattacked team in every setting, and in most settings it also brings the false data back to that level. The exceptions are two colluding attackers, fakes placed next to real drones, three of the ten caves and a team of 16 drones, and binding bodies to identities fixes the last three. Over 2{,}350 missions, an honest drone was revoked by a teammate only twice.
\end{abstract}


\section{\textbf{Introduction}}

\IEEEPARstart{S}{mall} drones are a natural fit for exploring buildings, mines and other places where a person cannot easily follow. A team covers such a space faster than a single drone, provided that the drones do not explore the same regions. Most coordinated exploration methods achieve this in two ways~\cite{burgard2005coordinated,zhou2022swarm}. Each drone announces the frontier it is heading for, so the others pick different ones. In addition, each drone shares the map it has built, so that no region is explored twice.

Both mechanisms rest on an implicit assumption: every identity on the network is a real drone, and the information it sends is true. A Sybil attack violates exactly this assumption~\cite{douceur2002sybil,newsome2004sybil}. A single compromised drone communicates under several identities. When these fake teammates claim frontiers, the honest drones stay away from those regions, and when they send fabricated scans, the honest drones mark unexplored space as explored or plan around walls that are not there. No drone collides and no message is lost; the mission simply ends early, with part of the building never seen and with false walls in every honest map.

Signing every message would stop new identities from appearing. However, it requires keys and a trusted authority before the mission starts, and it does nothing about a compromised drone that lies under its own, valid name. So we look at defences built only from what an exploring drone already has: its own sensors and its own map.

This paper makes three contributions. The first is a threat model for Sybil attacks on coordinated 3D exploration, with five attackers of increasing skill, one of which sends its fakes to the openings between floors and seals them with fake walls. The second is a series of five defence levels that use only onboard sensing, ending with a rule that binds each observed body to a single identity. The third is an evaluation on 70 maps of seven kinds, which shows how much the damage depends on the shape of the space, why auditing map content alone is not enough, and where the stronger defences still fall short.

Section~\ref{sec:related} covers related work. Section~\ref{sec:system} describes the exploration system, Section~\ref{sec:threat} the attacks and Section~\ref{sec:defence} the defences. Results follow in Section~\ref{sec:results}, and Section~\ref{sec:conclusion} concludes the paper with its limitations.

\section{\textbf{Related Work}}\label{sec:related}

\textbf{Multi-robot exploration and mapping:} Frontier-based exploration was introduced by Yamauchi~\cite{yamauchi1997frontier}, and Burgard et al.~\cite{burgard2005coordinated} extended it to teams. It is still the basis of most multi-robot exploration systems, recent aerial swarms included~\cite{zhou2022swarm}. The maps are usually probabilistic occupancy grids~\cite{hornung2013octomap}. Robust multi-robot mapping methods reject measurements and loop closures that disagree with each other~\cite{mangelson2018pcm,lajoie2020door,tian2022kimera}, which works well against perceptual aliasing and sensor outliers. These methods were not designed for an adversary who controls several identities and deliberately makes its data look consistent.

\textbf{Sybil and Byzantine robots:} The Sybil attack was first described for peer-to-peer systems~\cite{douceur2002sybil}, and Newsome et al.~\cite{newsome2004sybil} studied it in sensor networks. We follow their distinction between identities that act at the same time and identities that appear one after another. Gil et al.~\cite{gil2017spoof} detect spoofed robots from the physical signature of their wireless signals. The closest work to ours is IsoRank~\cite{sagwal2026isorank}, which finds Sybil identities in multi-robot exploration from the geometry of their broadcast positions; it shares two authors with this paper. Our evidence is different: we compare the map data and claims an identity sends with what the receiving drone sees for itself. The two approaches use different signals and could be combined, but we have not run IsoRank on our missions, so this paper makes no quantitative comparison with it. For robot swarms, Strobel et al.~\cite{strobel2018byzantine} limit Byzantine robots through blockchain voting, Wardega et al.~\cite{wardega2023blocklist} build a decentralized blocklist from accusations based on observed behaviour, and Luo et al.~\cite{luo2025robotobchain} store neighbour observations on a distributed ledger. Our presence check is a simpler form of neighbour observation that requires no ledger. A large body of work makes multi-robot consensus resilient to malicious agents, either by discarding extreme values received from neighbours~\cite{saldana2017resilient} or by weighting neighbours with trust values extracted from their wireless signals~\cite{yemini2022trust,cavorsi2023trust}. These methods protect a scalar or low-dimensional quantity that the team agrees on. A shared 3D map is far richer, and every honest drone can check part of it with its own sensors, which is the property our audit relies on.

\textbf{Trust models:} We use the Beta reputation system of J{\o}sang and Ismail~\cite{josang2002beta}. The evidence is not ratings but first-hand contradictions between what a drone was told and what it later saw.

\section{\textbf{System Model}}\label{sec:system}

\begin{figure}[t]
\centering
\includegraphics[width=\columnwidth]{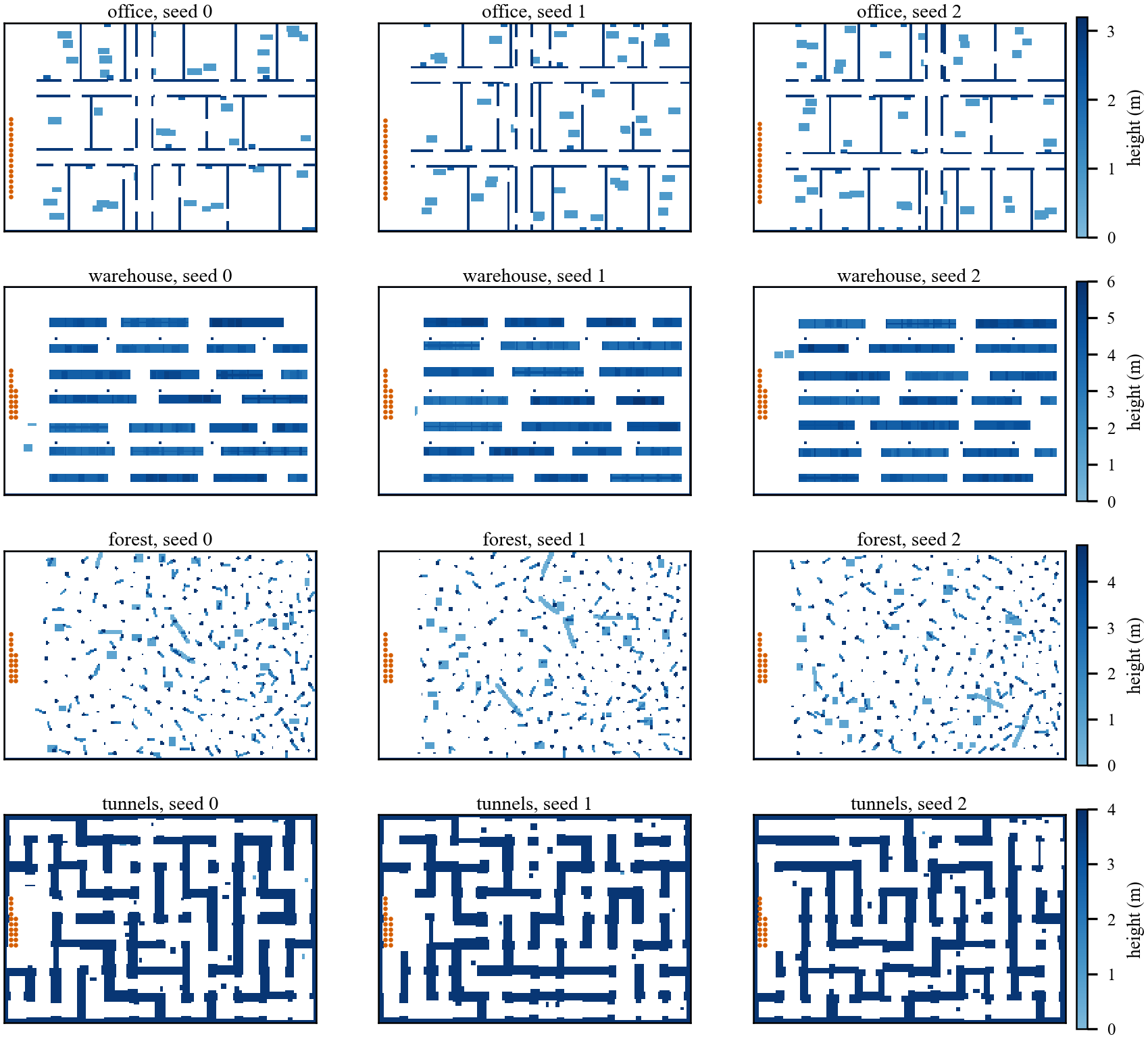}
\caption{Top views of three seeds of each world family. From top to bottom: office, warehouse, forest and tunnels. Obstacles that reach between 0.2 and 2.2\,m above the ground are coloured by their height, and the orange dots on the left mark the launch points.}
\label{fig:worlds}
\end{figure}

\begin{figure*}[t]
\centering
\includegraphics[width=\textwidth]{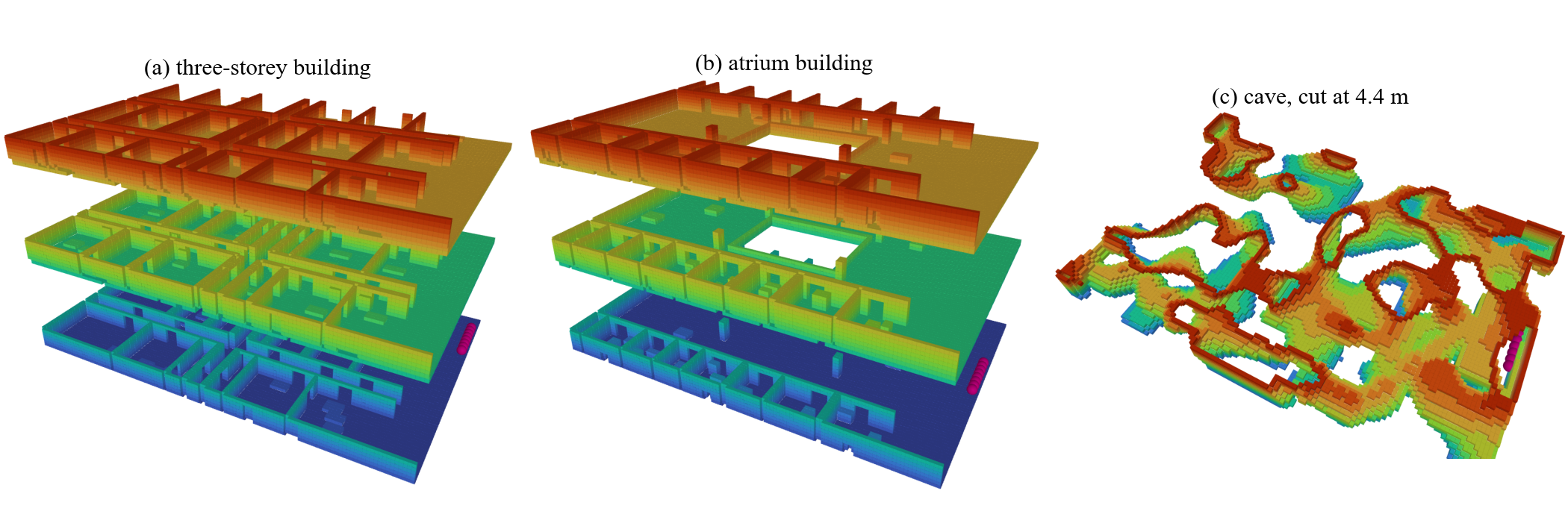}
\caption{The three extra world families (seed 0), rendered as 0.4\,m blocks coloured by height from blue (low) to red (high); roofs and the two near outer walls are removed. (a) Three-storey building and (b) atrium building, with the floors pulled apart vertically to show the shafts and the central void that connect them. (c) Cave, showing only rock below 4.4\,m, so the passages are visible from above. The magenta dots are the launch points.}
\label{fig:worlds3d}
\end{figure*}

\textbf{Worlds:} The main study uses four procedural families of $48\times32$\,m worlds with 0.2\,m voxels (Fig.~\ref{fig:worlds}): offices with rooms, corridors and furniture (3.2\,m high), warehouses with shelves (6\,m), forests with trunks and canopy (4.8\,m), and tunnel networks (4\,m). Seeds 0--9 of each family give the 40 maps of the main study. Seeds 100--104 are kept aside for calibration and never appear in the results. Three more families test vertical and confined exploration (Fig.~\ref{fig:worlds3d}). Three-storey buildings stack three office floors to 9.6\,m, joined by two 2\,m shafts through each slab. Atrium buildings have three floors of 3.6\,m around a central void with galleries and shops. Caves are 8\,m high, carved from smoothed random noise around one connected passage. We use ten seeds of each.

\textbf{Drones:} Each of the eight drones carries a forward-looking depth camera ($90^\circ\times60^\circ$, 6\,m range, 1.5\% range noise, 2\% dropouts) and two range sensors pointing up and down. Top speed is 1.5\,m/s. Every drone keeps its own voxel map in log-odds form, picks frontiers by their size per distance on a 0.4\,m planning grid, plans with breadth-first search and follows the path by pure pursuit, with a safety guard. They take off one second apart. We assume each drone knows its own pose. In the main study a drone moves as a point that reaches the commanded velocity within the 3\,m/s$^2$ acceleration limit; Section~\ref{sec:results} repeats the key conditions with quadrotor dynamics.

\textbf{Communication:} Two drones can communicate within 25\,m, and every wall between them halves that range. A drone sends its map updates to its neighbours under a bandwidth limit and resends lost packets. It also broadcasts a short message with its identity and position, and a claim with the frontier it is heading for. A drone will not pick a frontier that another identity claimed within 6\,m during the last 15\,s~\cite{burgard2005coordinated}. Each map cell remembers which identity reported it, so it can be checked or rolled back later.

\textbf{Identities:} An identity is a number carried in each message. Messages are not signed, so any drone can start sending under a new number. The team is open, meaning that drones may join or leave during the mission, so no drone knows how many teammates to expect and a Sybil attack cannot be caught by counting. In our experiments, all honest drones take off at the start and stay for the whole mission; only the fake identities join later. We do assume that an identity already used by an honest drone cannot be taken over, for instance because that drone would receive messages carrying its own number. Impersonation is outside the scope of this paper.

\section{\textbf{Threat Model}}\label{sec:threat}

One drone in the team is compromised (two in the collusion test). It explores and shares its own map honestly, so the team keeps trusting its real identity. We therefore do not test a compromised drone that lies under its own identity; D1 would audit such a drone like any other identity, but we leave this case to future work. In addition, it poses as up to $k$ additional drones at a time. These fake identities have no body; they exist only as messages.

\textbf{What the attacker knows and controls:} The attacker has its own map and receives every message sent within its radio range: map updates, position messages and frontier claims. We also assume it knows exactly how the defences work. It does not know what the other drones have seen but not yet shared. For each fake it decides three things: where the fake says it is, which frontier the fake claims, and what the fake reports seeing. It derives all three from its own map and from the messages it has received. We consider five attackers of increasing sophistication.

\textbf{Weak:} All $k$ fakes appear at the start, close to the attacker, and move openly. Their scans are random, with walls reported at random distances. This attacker is the easiest to detect, but it causes considerable damage before it is revoked.

\textbf{Strong:} This attacker is more careful. Its fakes join one at a time, the first after 8\,s and then one every 4\,s, each within 10\,m of the attacker. Instead of random walls, each fake reports a plausible room around itself, 8 to 14\,m across. The attacker also relabels 80\% of the real map data it forwards with the names of its fakes, so the fakes look honest most of the time. Once most of the team has revoked a fake, the attacker replaces it with a new one.

\textbf{Stealth:} The same as the strong attacker, except that the fakes only go to places that no honest drone can see at that moment. Consequently, no drone can ever observe that the claimed position is empty.

\textbf{Targeted:} The same as the strong attacker, except that the fakes are directed to the most valuable frontiers. Each fake picks the frontier with the most unexplored space behind it, such as a shaft leading to another floor or a doorway into a large unexplored area, and claims it. On arrival, it reports a wall across the opening, about 1\,m beyond the frontier. To the honest drones it then appears that nothing remains to be explored behind that opening. The attacker picks these spots from its own map, so the same rule applies in every world family.

\textbf{Shadow:} The same as the targeted attacker, except that each fake reports a position 0.3\,m from the position reported by a real drone. Any drone that checks this position observes a body where the fake claims to be. The body belongs to someone else, but the check cannot tell. This attacker is designed to defeat the presence check described below. The 0.3\,m offset is fixed; an adaptive attacker could instead place a fake at, or slightly ahead of, a real drone's broadcast position, which we do not test.

\section{\textbf{Defences}}\label{sec:defence}

The defences come in five levels. Each level keeps everything from the level before it and adds one new rule.

\textbf{First-hand audit (D1):} Each drone remembers who told it about every cell in its map. Whenever the drone later sees one of those cells with its own sensors, it checks whether the cell matches what it was told. For each identity $o$ it counts how many of $o$'s cells it has checked, $n_o$, and how many turned out wrong, $c_o$. A single mismatch can be innocent, since sensors are noisy and walls have edges, so a mismatch only counts as wrong if (i) within two voxels of the cell, neither the drone itself nor an identity it already trusts (see D2) has reported the same value as $o$, and (ii) the drone has seen at least 70\% of the area around the cell itself. Once per second, the drone asks how sure it is that more than 2\% of what $o$ told it is wrong. To answer this, it treats the true error rate of $o$ as unknown and describes it with a Beta distribution, the standard way to estimate a rate from a count of hits and misses. Starting from a uniform prior, where every rate is equally likely, $c_o$ wrong cells out of $n_o$ give $\text{Beta}(1+c_o,\,1+n_o-c_o)$~\cite{josang2002beta}. Few checks give a wide distribution and many checks a narrow one, so a drone does not revoke an identity on thin evidence. If $P(\text{error rate} > 2\%) > 0.99$, the drone \emph{revokes} $o$: it stops trusting $o$, deletes everything $o$ told it and asks its neighbours to send their maps again. The 2\% threshold comes from the calibration maps, where the worst honest drone reached only 0.04\% (0.43\% on wall cells alone).

\textbf{Vouching and presence (D2):} A drone starts to trust, or \emph{vouch for}, an identity once it has checked at least 2{,}000 of its cells and found at most 1\% of them wrong, half the revocation threshold. Only trusted identities can back up a received cell in test~(i). Drones also check the positions that other identities report. If a claimed position is inside the camera's view with nothing in the way, the drone looks for a drone body within $0.4\,\text{m} + v_\text{max}\,\Delta t$ of it, where $\Delta t$ is the age of the position message. A body that is really there is found with probability 0.9. Let $a_o$ count the checks with no body and $s_o$ the checks with one. If the drone is more than 99\% sure that the body is missing more than half of the time, that is, $P(\text{absence rate} > 0.5) > 0.99$ under $\text{Beta}(1+a_o,\,1+s_o)$, it revokes the identity.

\textbf{Physical evidence (D3):} An identity can only be trusted after its body has been seen at least once ($s_o > 0$). Walls get their own check with the same test, because a handful of fake walls can block a route while thousands of correct empty cells hide them in the overall error rate. Areas reported only by identities the drone does not trust yet are visited and checked, but with lower priority when the drone picks where to go next.

\textbf{Quarantine (D4):} At this level a drone ignores the maps and claims of every identity it has not physically confirmed. Confirmed means that the drone has seen the body itself, or that at least two other drones whose bodies it has seen say they have seen it; this two-witness rule is the \emph{quorum}. These witness reports are sent along with the position messages. Data from unconfirmed identities is kept, but not used until they are confirmed. Note what confirmed means here: a body was seen where the identity said it would be. That does not prove that this body sent the messages, and this is exactly the gap the shadow attacker exploits.

\textbf{One body, one identity (D5):} The presence check only tells a drone that \emph{some} body is at a position, not \emph{whose} body it is. D5 narrows this gap by linking each body a drone sees to exactly one identity. When several identities claim to be near the same body, the one whose reported position is closest wins. A real drone reports its own position, which sits right on its body, so it beats a fake placed 0.3\,m away. This margin is small: at 1.5\,m/s, a position message a few hundred milliseconds old is already about 0.3\,m out of date, and drift in a real pose estimate would shrink it further. Once a body is linked to an identity, any other identity that claims that body counts as missing, which quickly leads to its revocation. The links last for the rest of the mission. The same rule can therefore turn against an honest drone: if its body is ever bound to the wrong identity, the honest drone itself counts as missing from then on.

\textbf{Parameters:} Only the 2\% audit threshold was set on the calibration maps. The other values (2{,}000 checked cells, the 1\% vouching rate, the 0.9 detection rate, the 0.99 confidence levels and the quorum of two) were picked by hand while the system was being developed, on maps that may include the evaluation maps, and then frozen for every experiment in this paper.
For D5 we compared a binding that expires after 5\,s with one that never expires, against the shadow attacker on the five calibration offices (seeds 100--104). The binding that never expires left 45 to 61 false cells per honest map, against 118 to 3{,}253 with expiry, and neither revoked an honest drone, so we kept it.

\section{\textbf{Simulation Results}}\label{sec:results}

\textbf{Setup:} Missions last 600\,s (1{,}200\,s in the three-storey and atrium buildings) with eight drones unless stated otherwise. We measure three things: the \emph{explored space}, which is the share of reachable space that honest drones have seen with their own sensors; the \emph{false cells}, which is the median, over maps, of the number of cells per honest map that differ from the true world by more than one voxel; and the time until 90\% of the space is explored, averaged over the maps. The main study runs 36 conditions on each of the 40 maps (1{,}440 missions). The shadow attacker and D5 add 320 more, the team-size study another 320, and the three extra families 210, for 2{,}290 missions with point-mass motion; the dynamics check below adds 60 more. Each condition runs once on each map, so every number is taken across maps, not across repeated runs. We report the mean for the explored space and the median for the false cells, because the false cells are heavy-tailed and a few maps would dominate a mean. For every number in the tables, the repository also gives the range that contains the true value with 95\% confidence across maps (\texttt{results/intervals.md}); Table~\ref{tab:targeted} prints them for the ten-map experiments, where they matter most. Differences that lie within these ranges should not be read as real.

\begin{figure*}[t]
\centering
\includegraphics[width=\textwidth]{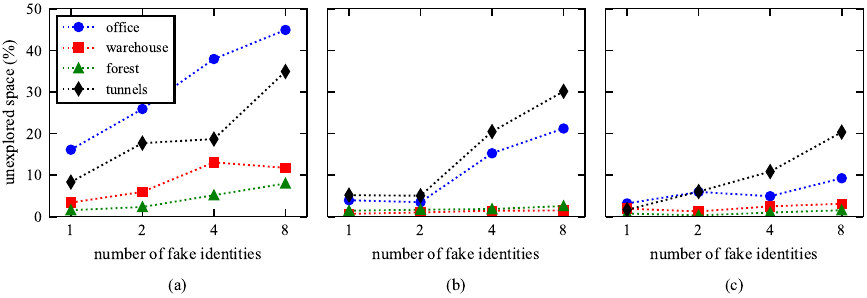}
\caption{Unexplored space without a defence against the number $k$ of fake identities run by one compromised drone, for each world family. (a) Weak attacker. (b) Strong attacker. (c) Stealth attacker. Each point is the mean over ten maps.}
\label{fig:damage}
\end{figure*}

\begin{figure*}[t]
\centering
\includegraphics[width=\textwidth]{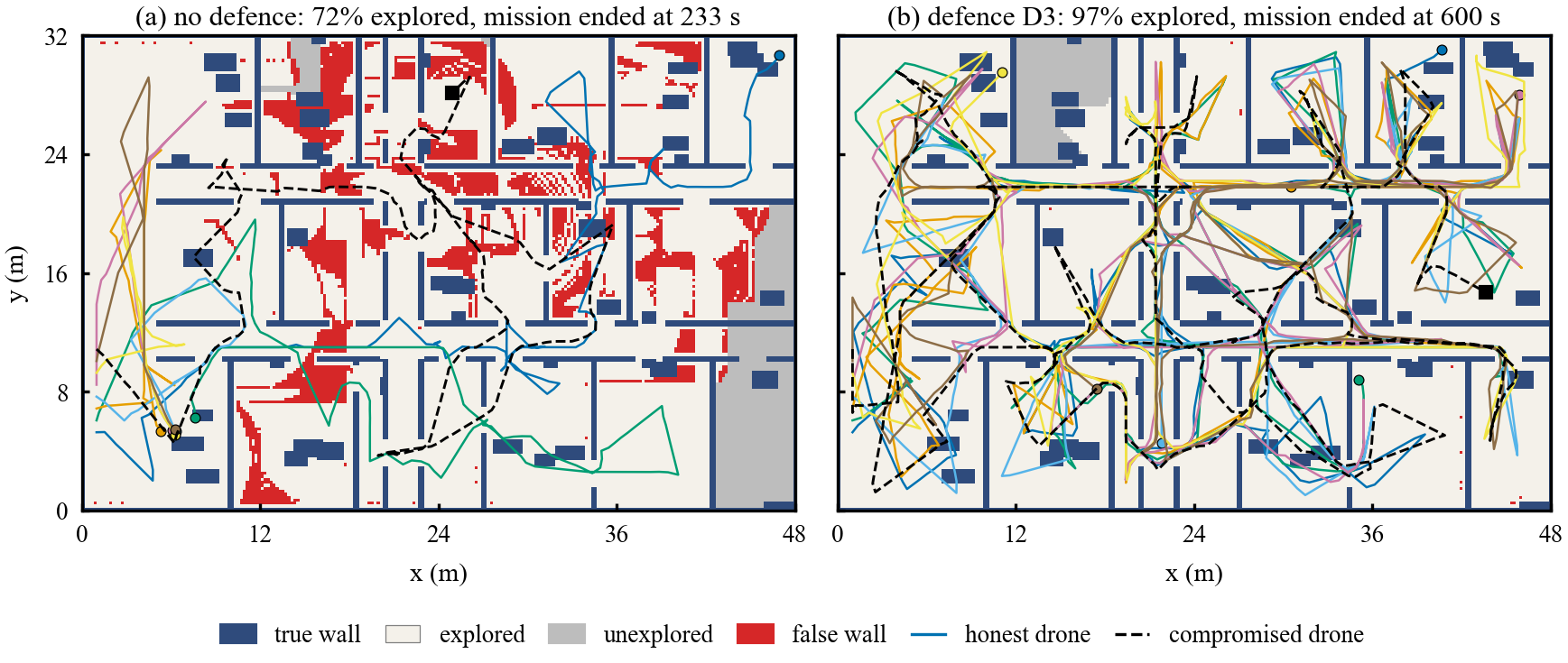}
\caption{Top view of one office mission (map 0) with the strong attacker and four fake identities, showing the merged map of the honest drones between 0.4 and 2.4\,m above the ground and each drone's path. (a) Without a defence, the team stops after 233\,s with 72\% of the office explored; the red cells are walls that do not exist. (b) With defence D3, the team keeps exploring until the end of the mission and covers 97\%.}
\label{fig:demo}
\end{figure*}

\begin{figure*}[t]
\centering
\includegraphics[width=\textwidth]{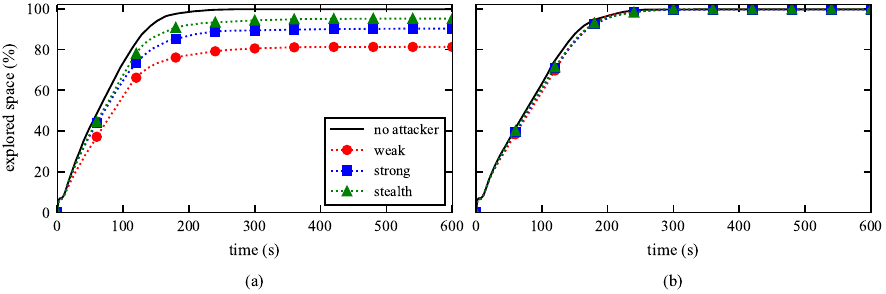}
\caption{Explored space over time with four fake identities. (a) Without a defence. (b) With the quarantine D4. Each curve is the mean over the 40 maps.}
\label{fig:time}
\end{figure*}

\textbf{Damage without a defence:} Fig.~\ref{fig:damage} plots the unexplored space against the number of fake identities, family by family, and Table~\ref{tab:damage} gives the averages. The type of environment has a strong effect. Offices and tunnels are split into rooms and branches reached through a few openings, and there eight weak fakes leave 45\% and 35\% of the space unexplored. Warehouses and forests are open and well connected. The same attack leaves only 12\% and 8\% there, and the strong and stealth attackers have little effect. At every $k$ the weak attacker does the most damage, while the stealth attacker trades some damage for being hard to check. Fig.~\ref{fig:time}(a) shows the mechanism: attacked teams explore almost as fast as honest ones at first, then stop early, because the rest of the space looks explored already. Fig.~\ref{fig:demo} shows one such mission in an office.

\begin{table}[t]
\centering
\caption{Explored space without a defence for different numbers of fake identities (mean over 40 maps). Without an attacker, the team explores 99.8\% of the space.}
\label{tab:damage}
\begin{tabular}{lcccc}
\toprule
attacker & $k=1$ & $k=2$ & $k=4$ & $k=8$ \\
\midrule
weak    & 92.7\% & 87.0\% & 81.3\% & 75.1\% \\
strong  & 97.2\% & 97.2\% & 90.3\% & 86.1\% \\
stealth & 98.1\% & 96.6\% & 95.2\% & 91.4\% \\
\bottomrule
\end{tabular}
\end{table}

\textbf{Effect of the defences:} The first-hand audit (D1) alone is sufficient to restore exploration. The explored space returns to 99.4--99.8\% (Table~\ref{tab:levels}), because fakes whose data contradicts what honest drones see get revoked, and their cells are deleted with them. Removing the false data is harder (Table~\ref{tab:family}). D1 removes the weak attacker's false data in every family, but the strong and stealth attackers are harder to remove. Their scans agree with themselves and are partly real, so D1 leaves 3{,}000 to 4{,}800 false cells per honest map, and 1{,}100 to 1{,}800 still survive D3. Tunnels are the most affected family, with about 40{,}000 false cells left after D3. D2 does not always improve on D1. Against the strong attacker it does, presumably because its presence check catches fakes that claim positions in view. Against the stealth attacker, whose fakes avoid every camera, the presence check never fires, and D2 leaves more false data than D1 in offices (on nine of the ten maps) and forests. Most likely this comes from vouching, D2's other addition: a stealth fake that relabels enough real data can earn trust, and its reports then back up the other fakes' cells in test~(i). D3, which trusts only identities whose body has been seen, removes this effect. Only the quarantine brings the false data down to the level of a mission without any attacker (56--63 cells, against 58--63). In tunnels, 116 to 135 false cells remain, against 107 without an attacker.

\input{table_family}

\begin{table}[t]
\centering
\caption{Explored space (\%) and false cells per honest map with four fake identities, for each defence level (40 maps).}
\label{tab:levels}
\setlength{\tabcolsep}{2.4pt}
\begin{tabular}{lccccc}
\toprule
attacker & none & D1 & D2 & D3 & D4 \\
\midrule
weak    & 81.3\,/\,24208 & 99.4\,/\,60   & 99.2\,/\,57   & 99.3\,/\,58   & 99.8\,/\,56 \\
strong  & 90.3\,/\,18726 & 99.7\,/\,4776 & 99.9\,/\,3465 & 99.7\,/\,1816 & 99.7\,/\,61 \\
stealth & 95.2\,/\,16234 & 99.8\,/\,3002 & 99.7\,/\,3686 & 99.8\,/\,1143 & 99.8\,/\,63 \\
\bottomrule
\end{tabular}
\end{table}

\textbf{Cost of the defences:} The defences come at a cost. With no attacker, reaching 90\% of a map takes 134\,s without a defence, 153\,s with D3 and 155\,s with D4, since drones spend time checking data they cannot vouch for yet. Under the strong and stealth attacks the order is reversed: D4 gets there in 164\,s and D3 in 213\,s. D3 keeps sending drones to verify what the fakes report. D4 does not use those reports at all.

\textbf{Collusion:} Against two compromised drones running four fakes each, D4 still brings exploration back to 99.7\%, but 1{,}386 false cells per honest map remain (Table~\ref{tab:collusion}). Most likely each attacker vouches for the other's fakes, which is enough to meet the quorum of two; we have not isolated this mechanism.

\begin{table}[t]
\centering
\caption{Explored space (\%) and false cells per honest map with eight fake identities run by one drone, and with four identities run by each of two colluding drones (40 maps).}
\label{tab:collusion}
\begin{tabular}{lccc}
\toprule
attacker & none & D3 & D4 \\
\midrule
no attacker & 99.8\,/\,62    & 99.7\,/\,63   & 99.7\,/\,58 \\
weak, $k=8$     & 75.1\,/\,30828 & 99.4\,/\,68   & 99.5\,/\,58 \\
strong, $k=8$   & 86.1\,/\,26661 & 99.8\,/\,1618 & 99.5\,/\,64 \\
stealth, $k=8$  & 91.4\,/\,24058 & 99.7\,/\,1373 & 99.8\,/\,61 \\
two attackers, $k=4$ & 76.3\,/\,35486 & 99.7\,/\,1831 & 99.7\,/\,1386 \\
\bottomrule
\end{tabular}
\end{table}

\textbf{Fakes next to real drones:} The shadow attacker places each fake 0.3\,m from a real drone, so the presence check finds a body where the fake says it is. This is sufficient to defeat D4 (Table~\ref{tab:bind}). With no defence the shadow attacker leaves 17{,}098 false cells per honest map; D3 cuts this to 1{,}782 and D4 only to 782, while the strong attacker is held to 61 by D4. In tunnels 38{,}528 false cells survive D4. D5 closes this gap for this attacker. A fake that claims a body already bound to a real drone counts as absent and is revoked shortly afterwards, and the median drops to 57, the same as an honest team under D5 (58). D5 also keeps the strong, stealth and targeted attackers at the honest level (59--60 cells). It is even faster than D4 under the shadow attack, reaching 90\% after 163\,s instead of 204\,s on average, because the fakes are revoked early and the drones no longer spend time checking their reports.

\begin{table}[t]
\centering
\caption{Shadow attacker (fakes placed next to real drones) and defence D5 (one body, one identity), four fake identities, 40 maps: explored space (\%), median false cells per honest map, and the median in tunnels.}
\label{tab:bind}
\setlength{\tabcolsep}{4pt}
\begin{tabular}{llccr}
\toprule
attacker & defence & explored & false cells & tunnels \\
\midrule
none     & D5   & 99.7 & 58     & 119 \\
shadow   & none & 95.7 & 17{,}098 & 139{,}099 \\
shadow   & D3   & 99.8 & 1{,}782 & 42{,}671 \\
shadow   & D4   & 99.8 & 782    & 38{,}528 \\
shadow   & D5   & 99.8 & 57     & 111 \\
strong   & D5   & 99.7 & 60     & 114 \\
stealth  & D5   & 99.7 & 60     & 111 \\
targeted & D5   & 99.6 & 59     & 117 \\
\bottomrule
\end{tabular}
\end{table}

\textbf{Team size:} Table~\ref{tab:teamsize} repeats the strong attacker in teams of 4 and 16 drones. Smaller teams are affected more. With 4 drones the explored space falls to 86.0\%, against 90.3\% with 8 and 98.6\% with 16, because a larger team has enough real drones to reach the regions the fakes claim. The false data, however, stays high at every size (12{,}413 to 20{,}632 cells). D4 and D5 both work with 4 drones. With 16, D4 leaves 121 false cells against 65 for the honest team, most likely because more bodies are around for fakes to be matched to by mistake. D5 stays at the honest level (61).

\begin{table}[t]
\centering
\caption{Strong attacker with four fake identities in teams of 4, 8 and 16 drones (40 maps): explored space (\%) and median false cells per honest map. The 8-drone values are from Tables~\ref{tab:levels}--\ref{tab:bind}.}
\label{tab:teamsize}
\setlength{\tabcolsep}{3pt}
\begin{tabular}{lcccc}
\toprule
drones & no attacker & strong & strong, D4 & strong, D5 \\
\midrule
4  & 99.7\,/\,48 & 86.0\,/\,20632 & 99.6\,/\,43  & 99.6\,/\,43 \\
8  & 99.8\,/\,62 & 90.3\,/\,18726 & 99.7\,/\,61  & 99.7\,/\,60 \\
16 & 99.8\,/\,65 & 98.6\,/\,12413 & 99.8\,/\,121 & 99.7\,/\,61 \\
\bottomrule
\end{tabular}
\end{table}

\textbf{Three-storey buildings:} Table~\ref{tab:targeted} sums up the three extra families. In the three-storey buildings (ten maps, 70 missions), the team explores 99.8\% of the building without an attacker. The strong attacker cuts this to 85.5\% on average. On most maps the effect is small, but on two maps whole floors are cut off (61.9\% and 34.4\% explored). The targeted attacker does damage more reliably. The median explored space drops from 96.3\% under the strong attacker to 83.4\%, the team stays below 92\% on seven of the ten maps instead of three, and on one map the team hardly progresses beyond the ground floor (34.6\%). The targeted attacker also adds the most false data in these buildings, about 72{,}000 cells per honest map. D3 and D4 both bring the explored space back to 99.8\% on average. With D4 the median number of false cells (189) matches the honest team (193), while 453 remain with D3. As in the main study, D4 is faster than D3 under attack: 90\% of the building is explored after 466\,s with D4 and 658\,s with D3, against 377\,s for the honest team with no defence and 418\,s for the honest team with D4.

\textbf{Atrium buildings and caves:} In the atrium buildings the floors meet around a large central void instead of narrow shafts, so there is no small opening for the attacker to seal. Both attackers therefore do less damage than in the three-storey buildings. The explored space falls to 94.1\% under the strong attacker and 93.1\% under the targeted attacker, and the worst map keeps 85.3\%. The false data is still of the same order (46{,}000 to 53{,}000 cells). D3 brings the explored space back to 99.7\% but leaves 2{,}346 false cells per honest map, against 166 for the honest team, while D4 leaves 163. The caves are the hardest environment. Even without an attacker, the team explores less than 92\% of three of the ten caves within the 600\,s mission. Under attack the mean explored space falls to 82.9\% and 86.5\%, with 11.7\% on the worst map, and the false data reaches 257{,}000 and 348{,}000 cells per honest map, more than in any other family. As in the tunnels, D3 brings the explored space back to the level of the honest team (93.6\% against 93.4\%) but leaves most of the false data in place (168{,}751 cells). D4 removes it on seven of the ten caves, where the false cells drop to the honest level, and the median is 45 against 34 for the honest team. On the other three caves, however, 993, 5{,}070 and 13{,}871 false cells survive D4. Most likely, fakes in these tight passages are confirmed against nearby real bodies, the same weakness the shadow attacker exploits, and D5 indeed fixes all three (36, 29 and 33 cells). With D5 the median over the caves is 33, against 34 for the honest team, and in the three-storey and atrium buildings D5 matches the honest team as well (184 and 160 cells).

\begin{table*}[t]
\centering
\caption{Three-storey buildings, atrium buildings and caves with four fake identities. Mean explored space (\%) and median false cells per honest map, each with the range that contains it with 95\% confidence over the ten maps of the family (k = thousand cells).}
\label{tab:targeted}
\footnotesize
\setlength{\tabcolsep}{2.6pt}
\begin{tabular}{llcccccc}
\toprule
 & & \multicolumn{2}{c}{three-storey} & \multicolumn{2}{c}{atrium} & \multicolumn{2}{c}{cave} \\
\cmidrule(lr){3-4}\cmidrule(lr){5-6}\cmidrule(lr){7-8}
attacker & defence & explored & false cells & explored & false cells & explored & false cells \\
\midrule
none & none & 99.8 (99.6--100.0) & 193 (170--201) & 99.8 (99.7--99.9) & 166 (158--175) & 93.4 (86.8--98.4) & 34 (27--41) \\
none & D4 & 99.8 (99.5--100.0) & 178 (170--192) & 99.8 (99.6--99.9) & 169 (157--188) & 93.4 (86.8--98.5) & 38 (30--44) \\
strong & none & 85.5 (71.6--96.0) & 44.6k (25.4k--58.8k) & 94.1 (92.4--96.0) & 45.8k (37.4k--50.3k) & 82.9 (65.1--95.0) & 257.1k (184.9k--329.1k) \\
targeted & none & 82.3 (70.5--91.3) & 72.1k (42.3k--82.3k) & 93.1 (90.8--95.2) & 52.8k (45.6k--69.1k) & 86.5 (75.9--95.1) & 347.8k (294.5k--456.1k) \\
targeted & D3 & 99.8 (99.5--100.0) & 453 (328--1{,}826) & 99.7 (99.5--99.9) & 2{,}346 (1{,}569--6{,}351) & 93.6 (87.0--98.7) & 168.8k (138.8k--221.0k) \\
targeted & D4 & 99.8 (99.5--100.0) & 189 (177--196) & 99.8 (99.7--99.9) & 163 (156--169) & 93.4 (86.8--98.6) & 45 (30--2{,}565) \\
targeted & D5 & 99.8 (99.4--100.0) & 184 (167--201) & 99.8 (99.7--100.0) & 160 (151--169) & 93.4 (86.8--98.7) & 33 (27--42) \\
\bottomrule
\end{tabular}
\end{table*}

\textbf{Flight dynamics:} To check that the results do not depend on the point-mass motion, we repeated five conditions on twelve maps (seeds 0--2 of each family, 60 missions) with each drone modelled as a rigid body with the size of a Crazyflie~2: 27\,g, thrust limited to 0.35\,N (a thrust-to-weight ratio of 1.3, below that of the real vehicle, so the model responds more slowly than a real Crazyflie), 20\,ms motor lag and rotor drag, flown by a geometric controller~\cite{lee2010geometric} at 500\,Hz that tracks the same guidance commands. The depth camera and the range sensors are fixed to the body, so scans tilt with the drone, up to 34$^\circ$ here. Table~\ref{tab:physics} compares the two on the same maps. With D4 and D5 the outcome is unchanged: the team explores all of the space, the false cells stay at the level of an honest team, and no honest drone is revoked. Without a defence the attack still adds about 20{,}000 false cells per honest map, but the explored space changes strongly from map to map in both models (on one office map 30\% kinematic and 94\% with dynamics, on another 72\% and 60\%), so the twelve-map means of 80.6\% and 93.1\% should not be read as a difference between the models. The dynamics did cause collisions: 8 drones crashed in 6 of the 60 missions, four of them in one tunnel map, after overshooting by 6--8\,cm in narrow passages. The safety guard brakes as if the drone stopped at once, and a margin that accounts for the response time of the vehicle is needed on real drones.

\begin{table}[t]
\centering
\caption{Point-mass motion against Crazyflie~2 dynamics on the same twelve maps: explored space (\%) / median false cells per honest map, and crashed drones over all twelve missions (point mass / dynamics).}
\label{tab:physics}
\setlength{\tabcolsep}{3pt}
\begin{tabular}{llccc}
\toprule
attacker & defence & point mass & dynamics & crashed (pm\,/\,dyn) \\
\midrule
none   & none & 99.7\,/\,63    & 100.0\,/\,67   & 0\,/\,1 \\
none   & D4   & 99.7\,/\,59    & 99.7\,/\,61    & 0\,/\,4 \\
strong & none & 80.6\,/\,16549 & 93.1\,/\,21691 & 0\,/\,1 \\
strong & D4   & 99.6\,/\,63    & 100.0\,/\,65   & 0\,/\,1 \\
strong & D5   & 99.7\,/\,62    & 100.0\,/\,60   & 0\,/\,1 \\
\bottomrule
\end{tabular}
\end{table}

\textbf{False accusations:} In the 2{,}290 point-mass missions no drone collided with an obstacle, and over all 2{,}350 missions an honest drone was revoked only twice, both times in a team of 16 and both times by a single teammate while its other 13 honest teammates kept trusting it. Once the wall check (D3, also active under D4) revoked a drone after 4 contradictions among only 45 checked wall cells, and once D5 bound a body to the wrong identity in a crowded tunnel, the failure mode described in Section~\ref{sec:defence}. Revocation is a local decision, so a mistaken teammate only drops that drone's data from its own map.

\section{\textbf{Conclusion}}\label{sec:conclusion}

This paper studied Sybil attacks on drone teams that explore unknown 3D spaces together, and defences against them that use nothing but onboard sensing. One compromised drone runs fake identities, and we considered five ways of using them: fakes that send random scans, fakes that build trust slowly, fakes that hide from every camera, fakes that seal the openings between floors, and fakes that hide right next to real drones. Against these we built five defences, each adding to the last. A first-hand audit compares received map data with what the drone sees itself and revokes an identity once a Beta test shows, with 99\% confidence, that its contradiction rate is above 2\%. A presence check looks for a body where an identity says it is. Physical evidence lets a drone trust only identities whose body it has seen. A quarantine uses data only from such confirmed identities. Finally, binding each body to one identity prevents a fake from claiming the body of a real drone.

Three findings stand out from the 2{,}350 missions. First, the layout decides how much damage a few fakes can do. In offices and tunnels, eight weak fakes leave 45\% and 35\% of the space unexplored, while open warehouses and forests lose only 12\% and 8\%, and the floor-sealing attacker cuts whole floors out of a three-storey building. Second, getting exploration back is easy, but getting rid of false data is not. The audit alone restores 99.4--99.8\% of the space, yet fakes whose scans agree with themselves leave 3{,}000 to 4{,}800 false cells per honest map after the audit and still 1{,}100 to 1{,}800 after D3, and about 40{,}000 in tunnels, until the quarantine sets aside all data from identities whose bodies have not been confirmed. Third, seeing a body is only useful if the body is tied to one identity. Fakes placed next to real drones push 782 false cells per map past the quarantine, and the body binding brings this back to 57, the level of an honest team; it also clears the three caves and the 16-drone team where the quarantine fell short.

None of these defences authenticates an identity. They check that its data agrees with what honest drones see and that a body shows up where it claims to be, so they complement signed messages rather than replace them, and they could also be combined with detectors that use other evidence, such as broadcast geometry~\cite{sagwal2026isorank} or radio signatures~\cite{gil2017spoof}, which we did not run on the same missions. The results come from simulation, with known poses, an idealized depth camera, mostly point-mass motion (with quadrotor dynamics the defences behave the same, but drones overshoot in narrow passages) and the assumption that a drone can detect other drones and distinguish them. The attacks are also limited. The compromised drone never lies under its own identity, the shadow attacker keeps a fixed offset instead of adapting to message delays, and honest drones never join or leave during a mission, although the team is designed to allow it. D5 has a cost of its own: its bindings are permanent, so a single wrong binding makes an honest drone look absent, as happened once in our missions. Within these limits, the main gap is collusion: two compromised drones, most likely by vouching for each other's fakes, meet the quorum of the quarantine. Our next steps are to detect teammates on real drones, for example with cameras or ultra-wideband ranging, to make the audit tolerate drift in the pose estimate, and to find quorum rules that resist collusion without slowing down honest teams.

\bibliographystyle{IEEEtran}
\bibliography{refs}

\end{document}

%% file: table_family.tex
\begin{table}[t]
\centering
\caption{Median false cells per honest map with four fake identities, by world family and defence level (ten maps each). The last column gives the honest team without a defence.}
\label{tab:family}
\setlength{\tabcolsep}{3pt}
\begin{tabular}{llrrrrrr}
\toprule
attacker & family & none & D1 & D2 & D3 & D4 & honest \\
\midrule
weak & office & 10{,}041 & 55 & 56 & 56 & 53 & 60 \\
 & warehouse & 29{,}038 & 59 & 52 & 52 & 59 & 61 \\
 & forest & 15{,}578 & 47 & 48 & 57 & 51 & 54 \\
 & tunnels & 126{,}040 & 115 & 121 & 129 & 135 & 107 \\
\midrule
strong & office & 17{,}496 & 2{,}445 & 1{,}152 & 164 & 53 & 60 \\
 & warehouse & 19{,}737 & 5{,}337 & 4{,}174 & 2{,}155 & 66 & 61 \\
 & forest & 12{,}906 & 992 & 922 & 323 & 50 & 54 \\
 & tunnels & 122{,}971 & 66{,}507 & 54{,}662 & 38{,}925 & 120 & 107 \\
\midrule
stealth & office & 11{,}232 & 935 & 2{,}168 & 206 & 65 & 60 \\
 & warehouse & 22{,}463 & 4{,}194 & 3{,}844 & 1{,}765 & 50 & 61 \\
 & forest & 9{,}460 & 528 & 819 & 293 & 56 & 54 \\
 & tunnels & 124{,}418 & 59{,}800 & 58{,}618 & 42{,}423 & 116 & 107 \\
\bottomrule
\end{tabular}
\end{table}